\documentclass{article}

\usepackage[english]{babel}

\usepackage[letterpaper,top=2cm,bottom=2cm,left=3cm,right=3cm,marginparwidth=1.75cm]{geometry}

\usepackage{hhline}
\usepackage{amsmath}
\usepackage{graphicx}
\usepackage[colorlinks=true, allcolors=blue]{hyperref}
\usepackage{multirow}
\title{Computational Models to Study Language Processing in the Human Brain: A Survey}

\author{
  Shaonan Wang \\
  Institute of Automation Chinese Academy of Science \\
  {shaonan.wang@nlpr.ia.ac.cn}  \\ 
  \and
  Jingyuan Sun \\ 
  KU Leuven \\
  jingyuan.sun@kuleuven.be \\
  \and 
  Yunhao Zhang \\
  Institute of Automation Chinese Academy of Science \\ 
  zhangyunhao2021@ia.ac.cn \\ 
  \and 
  Nan Lin\\
  Institute of Psychology Chinese Academy of Science\\
  linn@psych.ac.cn \\
  \and
  Marie-Francine Moens\\
  KU Leuven\\
  sien.moens@kuleuven.be \\ 
  \and
  Chengqing Zong \\
  Institute of Automation Chinese Academy of Science \\
  cqzong@nlpr.ia.ac.cn  \\
}

\begin{document}
\maketitle
\begin{abstract}

Despite differing from the human language processing mechanism in implementation and algorithms, current language models demonstrate remarkable human-like or surpassing language capabilities. Should computational language models be employed in studying the brain, and if so, when and how? To delve into this topic, this paper reviews efforts in using computational models for brain research, highlighting emerging trends. To ensure a fair comparison, the paper evaluates various computational models using consistent metrics on the same dataset. Our analysis reveals that no single model outperforms others on all datasets, underscoring the need for rich testing datasets and rigid experimental control to draw robust conclusions in studies involving computational models.

\end{abstract}

\section{Introduction}

Can computational models unravel the mysteries of language processing in the human brain? As computational language models advance, interdisciplinary research increasingly leverages them to study the brain, raising questions about their benefits and conditions of effectiveness.

Critics argue that substantial disparities between these models and the human brain make them inappropriate for direct brain mechanism studies. One major critique focuses on the limited scope of studies establishing correlations between brain activations and model representations \cite{geirhos2020beyond, guest2023logical}. For instance, Guest and Martin (2023) caution against using artificial neural networks (ANNs) to conclude the mind and brain, citing potential logical fallacies: it is inappropriate to assert that if the model predicts neural activity, then the model represents the neural system. Conversely, stating that if the model represents the neural system, it predicts neural activity is appropriate \cite{guest2023logical}. A second critique underscores differences in objective functions, learning rules, and architectures when comparing models with human language processing. In the vision domain, it questions the general approach modeling human object recognition by optimizing classification performance may be misguided for a theoretical reason, namely, the human visual system may not be optimized to classify images \cite{bowers2022deep,zador2019critique}. Similarly, concerns extend to the word prediction objective function in language processing \cite{huettig2016prediction}. The third critique argues that computational model findings lack novelty, often restating existing knowledge. According to Barsalou (2017)," Neural encoding research tells us little about the nature of this processing. While mapping concepts between Marr’s computational and implementation levels to support neural encoding and decoding, this approach ignores Marr’s algorithmic level, central for understanding the mechanisms that implement cognition. \cite{barsalou2017does}". 

Despite valid concerns, as George E. P. Box noted, "All models are wrong, but some are useful." Advanced computational language models, despite fundamental implementation differences, emulate human language abilities. Viewing them as potential frameworks for understanding brain mechanisms offers three key advantages. Firstly, computational models efficiently quantify cognitive metrics and identify neural correlates in language processing. Compared to human annotations, they are cost-effective for large dataset annotation and excel in handling complex metrics like syntactic complexity. Utilizing these models for brain correlation provides greater flexibility in analyzing naturalistic data, while traditional contrasting methods are mainly used in controlled experiments \cite{hale2015modeling, wehbe2014aligning, toneva2022combining, reddy2021can, schrimpf2020integrative, zhang2022probing, sun2020neural}. Secondly, computational models, especially large language models, demonstrate human-like behavior in diverse language tasks, offering a way to piece together information from different modules and taking a holistic perspective to delve into brain language processing mechanisms. Integrating fragmented knowledge and combining disciplines, as emphasized by Kriegeskorte and Douglas (2018), is crucial for gaining theoretical insights in brain-computational models \cite{kriegeskorte2018cognitive}. Thirdly, these models generate prospective hypotheses to validate the linguistic phenomena underlying the brain \cite{cichy2019deep, baroni2022proper, kanwisher2023using}. If a model mimics human performance only with a specific structure, it implies that this architecture may capture information explaining observed behavior in the brain. In support of this notion, Kanwisher et al. (2023) propose deep networks can answer "why" questions about the brain, indicating optimization for a task drives observed phenomena.

To conduct a thorough examination of the efficacy of computational models in studying language processing within the brain, this research delves into the distinctive contributions made by statistical language models (SLMs), shallow embedding models (SEMs), and large language models (LLMs) over time. The study aims to elucidate how these models uniquely advance brain investigation, exploring specific contexts and methodologies. In the forthcoming sections, Section 2 provides the terminology for different computational models and cognitive measures. In section 3, we delve into the three advantages offered by these models, reviewing existing work on these aspects, and presenting a fair comparison of these models using the same training dataset and evaluation metric. Section 4 concludes the study, summarizing key findings and implications.

\section{Computational models and cognitive metrics}
Over the last two decades, three primary computational models have been employed to investigate brain language processing \cite{hale2022neurocomputational, arana2023deep}. Each of these models possesses unique structures and language capabilities, serving as valuable tools for quantification and simulation in the study of the human brain. By quantifying cognitive metrics and linking them to the brain, these models offer a means of quantifying the complexities of language processing in the brain, thereby bridging the gap between language symbols and neural processes.

\subsection{Cognitive metrics}
In the context of language and the brain, cognitive metrics encompass measures of comprehension, information processing speed, working memory load, and various factors associated with how the brain interacts with and processes linguistic information. These metrics include reaction time, error rate, eye-tracking measures, surprisal, entropy, syntactic complexity, and semantic processing. Here, we highlight three of the most commonly utilized metrics:

\begin{itemize}

    \item \textbf{Probability-related Mertics} One of the most common and direct metrics for measuring cognitive load is Surprisal, which quantifies the uncertainty associated with the occurrence of a word.  Mathematically, it is defined as the negative logarithm of the probability of the word:\( S(x) = -\log P(x) \). Another commonly used metric that focuses on information gain is Entropy Reduction, representing the decrease in uncertainty when a specific word is introduced. Mathematically, it is defined as the difference in entropy \( H(x) \) before and after combining with the word \( x_{i+1} \): \(H(x) - H(x | x_{i+1})\) with \(H(X) = - \sum_{i} P(x_i) \cdot \log_2(P(x_i)) \).

    \item \textbf{Syntactic-Related Metric} Parsing strategies analyze sentence structure in natural language. Notable approaches include top-down, bottom-up, and left-corner parsing. Top-down begins with overall sentence structure, bottom-up starts with individual words, and left-corner combines both, starting with leftmost constituents and expanding the parse tree based on the input sentence.

    \item \textbf{Semantic and syntactic-Related Metric} Representations in language involve encoding the meaning and syntactic structure of words, phrases, sentences, and discourses. Quality at each level influences overall language understanding. Effective representations capture hierarchical structure and semantic relationships within and between linguistic elements.

\end{itemize}

\subsection{Statistical language models}
Statistical language models estimate word or phrase sequence probabilities by analyzing relationships in a corpus. They predict the next word based on preceding words and incorporate explicit structures like grammar and parsing strategies. Here, we present illustrations of two extensively employed models utilized in the investigation of brain language processing.

\begin{figure*}[h]
    \centering
    \includegraphics[scale=0.5]{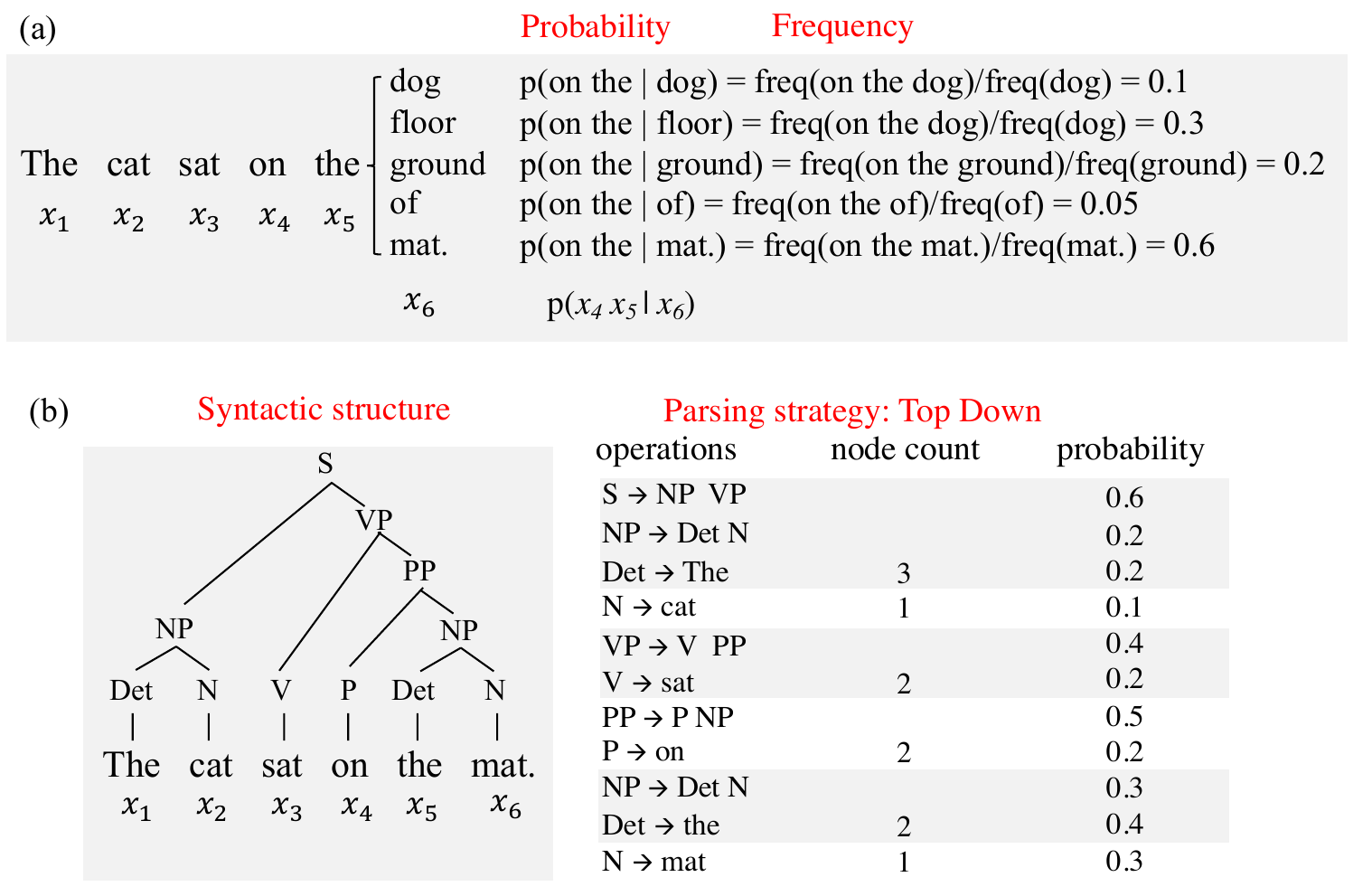}
    \caption{Statistical language models. (a) 3-gram language model, which estimates the word probability based on context. (b) structural language model, which incorporates syntax and parsing for understanding word interplay and sentence organization. }
    \label{fig:parse}
\end{figure*}

The n-gram language model estimates word probabilities based on word and n-gram frequencies. For example, with n=3, the probability of a word like ``mat'' in a sentence is calculated by analyzing the frequency of the preceding three words (``on the mat'') divided by the frequency of the target word itself (``mat''), as shown in Figure 1(a).

The structural language model utilizes formal grammars like Context-Free Grammar (CFG), Minimalist Grammar (MG), and Combinatory Categorial Grammar (CCG) to calculate word probabilities based on preceding words and syntactic structure of the sentence \cite{hale2001probabilistic}. For example, in Probabilistic Context-Free Grammar (PCFG), operation rules with associated probabilities are selected for non-terminal symbols like noun phrases and verb phrases (S $\to$ NP VP). As shown in Figure 1(b), to calculate a sentence's probability, first, operation rules for non-terminal symbols (e.g., S $\to$ NP VP) are chosen. These rules have individual probabilities, multiplied to compute the sentence's probability. This product signifies the overall likelihood of generating the sentence from the given grammar. Parsing strategies like top-down explore grammar rules from the root down, with probabilities guiding rule selection. In Figure 1(b), top-down parsing begins from the root, advancing downward through the tree. Operation rule selection relies on the probability of expanding non-terminals. The process yields node counts, indicating rules needed per word. For instance, ``sat'' requires two steps: VP$\to$P NP and P$\to$on.



\subsection{Shallow Embedding Models}
After statistical language models, pioneering shallow embedding models encode meaning through the distributional hypothesis, which posits that similar words occur in similar contexts. These models quantify semantics and transform linguistic entities into vector spaces \cite{pennington2014glove, le2014distributed, sun-etal-2018-memory}. 
 
Shallow embedding models, with diverse training data, design, and goals, share a common aim: capturing semantic nuances for improved language processing tasks. Notable examples include Word2Vec \cite{mikolov2013efficient} and the Recurrent Neural Network Language Model (RNNLM) \cite{mikolov2011extensions}, both exemplifying algorithms like Continuous Bag of Words (CBOW) and Skip-Gram within the Word2Vec model. Consider CBOW as illustrated in Figure 2(a). This model predicts a target word from its context. Mathematically, the algorithm is described as:
\[ L_{\text{CBOW}} = \frac{1}{T} \sum_{t=1}^{T} \log p(w_t | w_{t - c}, \dots, w_{t + c}) \]
RNNLM, as illustrated in Figure 2(b), on the other hand, leverages RNNs to predict the next word in a sequence, considering all the previous words. Its formulation can be given by:
\[ p(w_t | w_{1}, w_{2}, \dots, w_{t-1}) = \text{softmax}(h_{t}) \]
Where \( h_t \) is the hidden state at time \( t \), which is a function of the previous hidden state and the current input.

\begin{figure*}[h]
    \centering
    \includegraphics[scale=0.5]{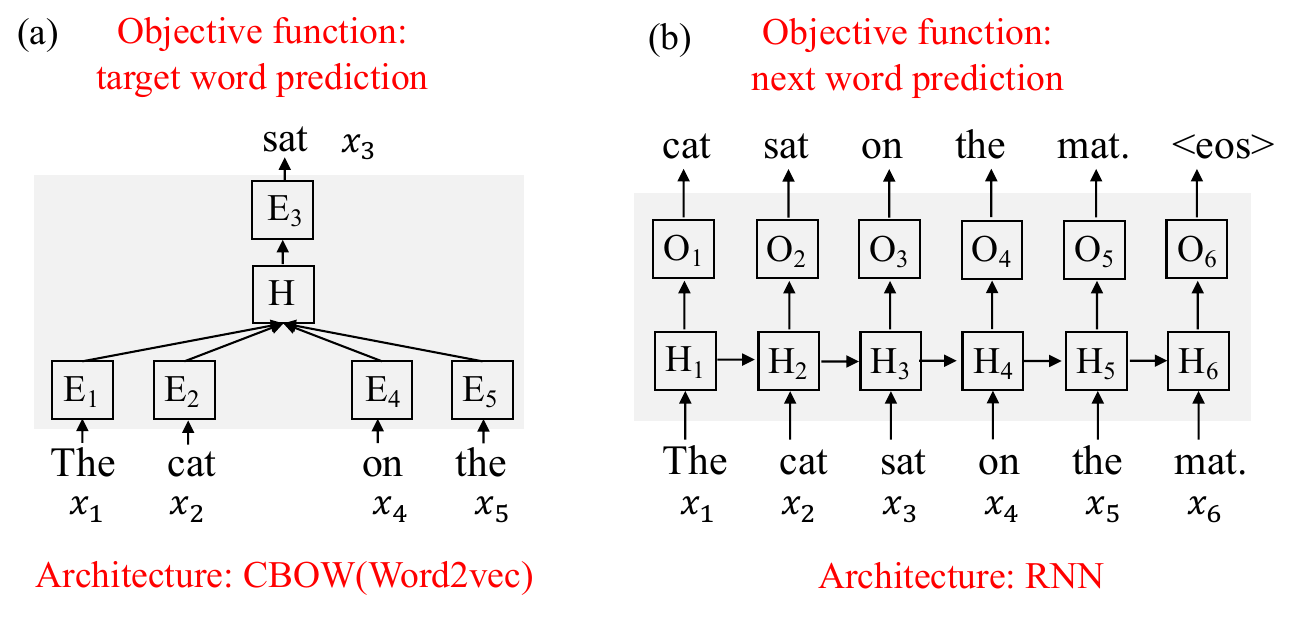}
    \caption{Shallow Embedding Models: (a) CBOW (Word2Vec) model that learn word embeddings by predicting target word based on context within a window, (b) RNN model that learning word embeddings by predicting next word based on all previous context.}
    \label{fig:parse}
\end{figure*}

Both models learn word representations through the utilization of hidden layers in neural networks. CBOW employs static word representations, while RNN acquires contextualized word representations. These representations capture the semantic and syntactic relations between words.

\subsection{Large Language Models}
Following the shallow embedding model era, large language models have revolutionized natural language processing, excelling in diverse tasks within a single model and quickly adapting to new tasks with minimal examples \cite{vaswani2017attention,devlin2018bert,radford2019language,liu2019roberta,touvron2023llama,kasneci2023chatgpt}. Their remarkable improvement in generating human-like text and surpassing human-like performance on various tasks may offer new insights for brain language studies.

GPT-3 and similar large language models \cite{brown2020language} distinguish themselves from shallow embedding models through extensive use of massive internet data, deep architectures with numerous parameters, and sophisticated algorithms for optimizing objective functions. This enables rapid in-context learning, facilitating swift adaptation to new tasks and providing human-like responses.

\begin{figure*}[h]
    \centering
    \includegraphics[scale=0.5]{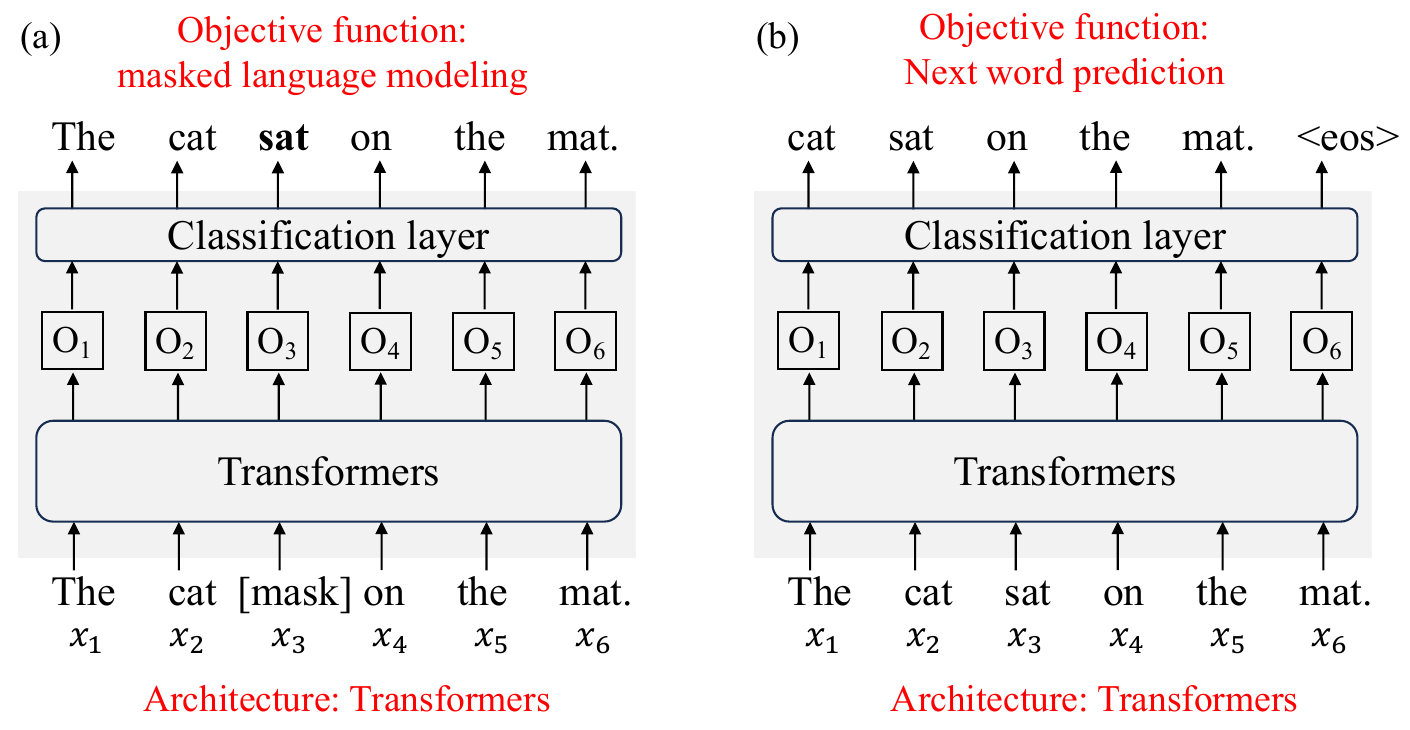}
    \caption{Large Language Models: (a) BERT: Learns word embeddings through masked language modeling, predicting randomly masked words based on surrounding context. (b) GPT: Learns word embeddings via next word prediction, predicting the next word based on all preceding context.}
    \label{fig:parse}
\end{figure*}

Large language models rely on transformers, employing multiple attention layers to capture intricate relationships in input data. This hierarchical architecture enables selective focus on different parts of the sequence, enhancing contextual understanding. A primary technique used in pretraining large language models is the masked language modeling (MLM), as shown in Figure 3(a). Here, certain words in a sentence are masked out, and the model is trained to predict the masked word based on its context. The objective for MLM can be represented as:
\[ L_{\text{MLM}} = - \log p(w_i | w_{\text{context}}) \]
where \( w_i \) is the masked word and \( w_{\text{context}} \) are the surrounding words. 
Another common training approach is next word prediction (NWP) as shown in Figure 3(b), where the model predicts the next word in a sequence given the preceding words. The objective for this is:
\[ L_{\text{NWP}} = - \log p(w_{t+1} | w_1, w_2, \dots, w_t) \]

Given the human-like performance and responses achieved by large language models, exploring the alignment of their objective functions, structural components, attention mechanisms, and encoded meanings with the human brain could yield intriguing insights.

\section{Utilizing computational models in brain studies}
This section reviews studies that employed computational models to quantify cognitive measures like surprisal, entropy, and semantic representations to study the neural mechanism underlying brain language processing. 

\subsection{Quantifying cognitive load}
Cognitive measures connect discrete language symbols with continuous neural processes in comprehension. For instance, higher surprisal or entropy reduction values signify increased cognitive load, demanding more effort for effective understanding. All computational models, including statistical, shallow embedding, and large language models, compute quantitative cognitive measures.

Prior research on statistical language models for surprisal consistently demonstrates effectiveness, particularly in eye-tracking studies, linking higher surprisal values to slower reading speeds, indicative of increased language complexity \cite{boston2008parsing, demberg2008data, rauzy2012robustness}. In event-related potentials, surprisal shows a positive association with the amplitude of the N400 component, supporting that surprisal as a generally applicable measure of processing difficulty during language comprehension \cite{frank2013word, frank2015erp}. Functional magnetic resonance imaging (fMRI) studies have demonstrated the predictive capacity of surprisal in determining activation patterns \cite{brennan2016abstract, hale2015modeling, henderson2016language}. Notably, Shain et al. (2020) observed that surprisal, derived from both structure-based and n-gram models, influences the language network but does not impact the domain-general multiple-demand network, indicating that predictive coding in the brain’s response to language is domain-specific and that these predictions are sensitive both to local word co-occurrence patterns and to hierarchical structure \cite{shain2020fmri}. Furthermore, Heilbron et al. (2022) employed GPT-2 to compute various types of surprisal, revealing that brain responses to words are modulated by pervasive predictions spanning from phonemes and syntactic categories (parts of speech) to semantics \cite{heilbron2022hierarchy}.

Prior research predominantly employs statistical language models to compute entropy reduction, which correlates with cognitive processing difficulty \cite{hale2003information, hale2006uncertainty}. This correlation holds across naturalistic text \cite{frank2013uncertainty, wu2010complexity} and controlled experiments \cite{linzen2016uncertainty}. Intracranial signals from the anterior Inferior Temporal Sulcus (aITS) and posterior Inferior Temporal Gyrus (pITG) also correlate with word-by-word Entropy Reduction values derived from phrase structure grammars for languages. In the anterior region, this correlation persists even when combined with surprisal co-predictors from PCFG and N-gram models, confirming that the brain's temporal lobe houses a parsing function. This function's incremental processing difficulty profile reflects changes in grammatical uncertainty \cite{nelson2017entropy}. A comprehensive analysis for a detailed comparison between surprisal and entropy reduction is provided by John Hale (2016) \cite{hale2016information}.


\subsubsection{Formalizing syntactic processing}

Parsing strategies like top-down, bottom-up, and left-corner in language processing employ unique syntactic analysis methods. It's still debatable whether syntax is encoded in the brain. Assuming it is, researchers compared word-by-word metrics from different parsing strategies with neural activity measurements to gain profound insights into the gradual formation of syntactic structures in the brain. Two frequently used metrics are rule counts and node counts. Prior research shows a correlation between parser-applied rules, node count (refer to Figure 1(b)), and related neural activity \cite{brennan2012syntactic, brennan2016abstract}. For example, comparing bottom-up and left-corner parsing reveals left-corner parse steps correlating with activity in the left anterior temporal lobe 350-500 ms after word onset \cite{brennan2017meg}. Moreover, continuous research emphasizes bottom-up and left-corner parsing's superiority in fitting activation patterns across the left-hemisphere language network over top-down parsing. These findings suggest that individuals might process simple sentence structures through bottom-up and/or left-corner parsing, with evidence favoring bottom-up parsing \cite{nelson2017neurophysiological}. In a comparative study by Zhang et al. \cite{zhang2022brain}, different languages' parsing strategies were explored, assessing working memory needs for varied language structures. The research showed that top-down parsing demands lower memory for right-branching English, while bottom-up parsing is less demanding for Chinese. Additionally, fMRI results indicated language-specific parsing preferences: Chinese favors bottom-up parsing, while English leans towards top-down parsing. For a comprehensive exploration of employing grammars and parsing strategies to investigate the construction of brain structures, consult the study by Brennan et al. (2016) \cite{brennan2016naturalistic}.

Comparing CCG, CFG, and LLM-based predictability, evidence suggests CCG captures neural activity beyond LLM and CFG parsing steps. Augmenting CCG with the reveal operation improves fits, particularly in right adjunction parsing. Strongest effects are observed in posterior temporal lobe, around the middle temporal gyrus \cite{stanojevic2023modeling}.

In terms of the syntactic structure of language processing, the structural language model derives surprisal estimates by analyzing the syntactic arrangement of a given text fragment. These estimates are then aligned with brain activation patterns to delve into the mechanisms underlying syntactic processing. For example, in comparing surprisals from sequential structure grammars and hierarchical phrase structure grammars, Frank and Bob (2011) found that the hierarchical sentence structure has minimal impact on anticipations for upcoming words \cite{frank2011insensitivity}. Conversely, Brennan et al. (2016) and Brennan and Hale (2019) found that predictions based on hierarchical structure, compared to sequential information, correlated more strongly with human brain responses in passive listeners engaged in an audiobook story, as evidenced by electroencephalography signals \cite{brennan2016abstract, brennan2019hierarchical}. The divergent conclusion suggests that studies based on models are sensitive to various factors, highlighting the need for standardization and uniformity. Further studies show that grammatical models such as CCG and RNNG, known for their expressiveness, provide a better match to neural signals than those derived from CFG. They notably capture activity in the left posterior temporal regions \cite{hale2018finding, brennan2020localizing, stanojevic2023modeling}. These findings advocate for CCG and RNNG as mechanistic models underpinning syntactic processing during standard human language comprehension.

Some studies explore full neural network architectures to understand human language processing. RNNs, CNNs, and Transformers are prominent in these discussions, with RNNs particularly used for their belief in the importance of recurrent processing in human language understanding.
Merkx and Frank (2021) compared Transformer-based and RNN-based language models in measuring human reading effort. Results indicate Transformers excel over RNNs in explaining self-paced reading times and neural activity during English sentence reading. This challenges the prevalent notion of immediate and recurrent processing, indicating a cue-based retrieval process in human sentence comprehension \cite{merkx-frank-2021-human}.
Seth et al. (2023) explore the relationship between CNN models and the brain, investigating if these models can explain object recognition solely based on visual properties without semantics. The study challenges the idea that semantic effects in the ventral visual pathway (VVP) during object recognition might be explained by higher-level visual object properties captured by CNN models\cite{seth_nicholls_tyler_clarke_2023}.

\subsection{Modelling linguistic Representations}

Recent research has leveraged text embeddings to explore language processing in the human brain, correlating brain activation from linguistic stimuli with corresponding stimulus embeddings \cite{Wehbe2014AligningCS, anderson2016representational, deHeer2017TheHC, Jain2018IncorporatingCI, sun2019towards, wang2020fine}. Techniques such as Representational Similarity Analysis and regression models have been pivotal in this analysis \cite{Wehbe2014SimultaneouslyUT, anderson2016representational, Huth2016NaturalSR,Anderson2019MultipleRO}. For example, Broderick et al. (2018) employed Word2Vec to measure semantic differences in narrative contexts, correlating these with EEG data to show EEG's reflection of semantic processing and its alignment with N400 characteristics \cite{broderick2018electrophysiological}. Similarly, Xu et al. (2016) connected shallow embeddings (GloVe, Word2Vec, RNN) with fMRI data from word observation tasks and found Skip-gram of Word2Vec and Glove performed quite well \cite{xu2016brainbench}.  Fu et al. (2023) compared word embeddings, co-occurrence, and graph-topological methods, highlighting their differing efficacies in semantic brain pattern mapping \cite{fu2023different}. Further, researchers have refined embeddings to better represent specific features, utilizing algebraic and neural network-based modifications \cite{Anderson2013OfWE, anderson2017visually, Lyu2019NeuralDO}. Studies like Zhang et al. (2020) and Toneva et al. (2022) used modified embeddings to explore brain-based word meaning relations, revealing complex patterns of semantic categories and relations \cite{mollica2020composition, toneva2022combining, zhang2020connecting}. Researchers have also extended embeddings to sentence-level representations, with average pooling being a popular method \cite{sun2019towards, pereira2018toward}.

Given the advanced prediction capabilities of large language models over shallow embeddings \cite{devlin2018bert, sun2020distill}, research now focuses on determining the most effective models for brain language representation \cite{oota2022deep, caucheteux2022deep, merlin2022language,balabin2023investigating}. Studies by Sun et al. (2021) compared multiple models, including large language models, confirming their superiority in predicting brain language network activities \cite{sun2020neural}. Antonello et al. (2021) introduced a novel "language representation embedding space" for predicting brain responses during language tasks \cite{antonello2021low}. Besides semantics, large language models capture syntactic and morphological aspects, with ongoing studies aiming to distinguish these features in brain processing \cite{antonello2021low, reddy2021can, oota2022long}. Caucheteux et al. (2021) \cite{caucheteux2021disentangling} developed an embedding taxonomy for large language models to analyze brain activities in narrative listening. They found compositional representations engaged a broader cortical network, including the bilateral temporal, parietal, and prefrontal cortices, more extensively than lexical representations.
Zhang et al. (2022) investigated the brain's syntactic processing using modified embeddings, revealing its distributed nature across brain networks \cite{zhang2022probing}.

Although large language model embeddings excel in aligning with brain activities, the underlying reasons and mechanisms of this synchrony are not fully understood \cite{lindborg2021meaning, pasquiou2022neural}. Research aims to clarify how these models mirror human linguistic comprehension and brain structures \cite{oota2022joint}. 
Sun et al. (2020) found that impairing semantic processing in large language models diminishes their brain activity alignment \cite{sun2020neural}. Conversely, Merlin and Toneva (2022) enhanced this alignment by focusing on next-word prediction and word-level semantics \cite{merlin2022language}. Aw and Toneva (2023) showed that training models on narrative summarization enhances brain activity synchronization \cite{aw2023training}. Caucheteux et al. (2023) found that integrating multi-timescale predictions into large language models improves brain mapping. Hierarchically, frontoparietal cortices predict higher-level, longer-range representations compared to temporal cortices, emphasizing the role of hierarchical predictive coding in language processing \cite{caucheteux2023evidence}. Further, task-specific tuning of large language models in NLP tasks has been shown to influence their brain pattern alignment \cite{sun2023ij, sun2023ec1, oota-etal-2022-neural}, indicating that their semantic feature capture is a key factor in this alignment.

\subsection{Verifying hypotheses for empirical validations}
Modern computational models, especially large language models, mimic human behavior and surpass human performance. They serve as valuable tools to verify and even generate hypotheses about the brain's language processing. Analyzing instances where task-optimized networks mirror human behavioral and neural patterns allows for the formulation of novel hypotheses about language processing in the brain \cite{kanwisher2023using}. 

Limited research has delved into this realm, with studies such as those by Schrimpf et al. (2021), Goldstein et al. (2022), and Caucheteux et al. (2022) found that Transformer-based large language models, optimized for next-word prediction, align well with both behavioral and neural data in humans. The stronger the model's performance in next-word prediction, the closer its match to human data, implying that prediction may be a key optimization aspect of the human language system \cite{schrimpf2021neural, goldstein2022shared, caucheteux2022brains}. Additionally, Zou et al. (2023) connected the human brain's attention in reading to the attention weights in Transformer models, suggesting a parallel mechanism between these models and the brain. This alignment implies that the brain's reading attention, like the Transformer models, is optimized for specific tasks \cite{Zou_2023}. These studies offer valuable insights into predictive coding and attention allocation hypotheses about language, yet only scratch the surface, confirming existing theories. 

Within the NLP domain, large language models are used to reveal the operational mechanisms of specific modules or neurons. Notably, Singh et al. (2023) demonstrated that these models can generate explanations for the response of individual fMRI voxels to language stimuli \cite{singh2023explaining}. Future research can explore using large language models to generate new theories or hypotheses, uncovering insights into their output behavior. This could motivate further studies on human brain language processing.

\section{Comparing models on a multimodal cognitive dataset}
Various models possess unique strengths. Among the three types of models, \textbf{SLMs (N-gram models)} offer simplicity and interpretability, while \textbf{structural models} leverage human-annotated trees for clear pattern-based predictions, requiring fewer computational resources than neural models. Despite advantages, statistical models have limited long-range dependency capture and sparse data issues for rare n-grams due to fixed context windows and word symbols. \textbf{SEMs} excel in semantic representation, addressing challenges like unreliable estimates and misspellings \cite{kwon2021handling,fu2014learning,edizel2019misspelling}. However, their static word representations lack the depth for intricate, context-dependent language processing. \textbf{LLMs} provide rich semantic insights with dynamic word representations \cite{naseem2020transformer}. Their depth captures intricate linguistic features and higher-order dependencies, facilitating swift in-context learning \cite{sengupta2020review,min2023recent,chan2022data,von2023transformers}. Yet, LLMs' complexity and black-box nature pose challenges in interpretation \cite{toneva2019interpreting,goldstein2022shared}.

Recent findings reveal that large language models demonstrate superior alignment with the brain compared to shallow embedding models, as evidenced by Schrimpf et al. \cite{schrimpf2021neural}. Nevertheless, challenges in conducting fair model comparisons persist due to variations in brain datasets and metrics.

To address this issue, we investigate how these models correlate with human cognitive data under fair comparison conditions using the same training data. We use English\footnote{https://dumps.wikimedia.org/enwiki/latest} and Chinese\footnote{http://www.xinhuanet.com/whxw.htm} datasets for training statistical (N-gram and structural models\footnote{https://nlp.stanford.edu/software/stanford-dependencies}), shallow (GloVe\footnote{https://nlp.stanford.edu/projects/glove/}, Word2Vec\footnote{https://radimrehurek.com/gensim/models/word2vec.html} with detailed parameters in \cite{emmanuele2021decoding}), and large (BERT-large\footnote{https://huggingface.co/bert-large-uncased}, GPT2\footnote{https://huggingface.co/gpt2-medium}, 24 layers, 1e-4 learning rate) language models to predict neuroimaging and eyetracking data. Word embeddings are extracted from layers yielding optimal performance\footnote{For word-level fMRI: BERT (en): 10th layer, GPT-2 (en): 1st layer, BERT (zh): 8th layer, GPT-2 (zh): 1st layer. Discourse-level fMRI: BERT (en): 13th layer, GPT-2 (en): 21st layer, BERT (zh): 23rd layer, GPT-2 (zh): 15th layer. Eye-tracking: BERT (en): 8th layer, GPT-2 (en): 4th layer, GPT-2 (zh): 3rd layer, BERT (zh): 19th layer.}). Models are trained from scratch\footnote{BERT and GPT2 were trained for one epoch, and consistent loss decrease was observed}. Testing on multi-modal cognitive datasets\cite{zhang2024mulcogbench}, encoding models predict fMRI or eye-tracking responses using 10-fold cross-validation. Paired t-tests assess group-level significance with a threshold of p=0.001.

\begin{table}[]
\begin{tabular}{|l|l|ccc|ccc|}
\hline
\multicolumn{1}{|c|}{}                                                       & \multicolumn{1}{c|}{}            & \multicolumn{3}{c|}{English}                                                                                                                                                                                                 & \multicolumn{3}{c|}{Chinese}                                                                                                                                                                                                 \\ \hline
\multicolumn{1}{|c|}{}                                                       & \multicolumn{1}{c|}{}            & \multicolumn{1}{c|}{\begin{tabular}[c]{@{}c@{}}fMRI\\ -word \\  \cite{pereira2018toward} \end{tabular}} & \multicolumn{1}{c|}{\begin{tabular}[c]{@{}c@{}}fMRI\\ -discourse \\ \cite{zhang2020connecting} \end{tabular} } & \begin{tabular}[c]{@{}c@{}}EyeTracking\\ -sentence \\ \cite{hollenstein2018zuco}\end{tabular} & \multicolumn{1}{c|}{\begin{tabular}[c]{@{}c@{}}fMRI\\ -word \\ \cite{wang2022fmri}\end{tabular}} & \multicolumn{1}{c|}{\begin{tabular}[c]{@{}c@{}}fMRI\\ -discourse\\ \cite{wang2022synchronized}\end{tabular}} & \begin{tabular}[c]{@{}c@{}}EyeTracking\\ -sentence\\ \cite{zhang2022database}\end{tabular}  \\ \hhline{|=|=|=|=|=|=|=|=|}
\multicolumn{1}{|c|}{\multirow{2}{*}{Surprisal}}                             & \multicolumn{1}{c|}{SLMs(Ngram)} & \multicolumn{1}{c|}{\textbackslash{}}                                     & \multicolumn{1}{c|}{0.014}                                                     & 0.178                                                           & \multicolumn{1}{c|}{\textbackslash{}}                                     & \multicolumn{1}{c|}{0.042}                                                     & 0.047                                                           \\ \cline{2-8} 
\multicolumn{1}{|c|}{}                                                       & \multicolumn{1}{c|}{LLMs(GPT2)}  & \multicolumn{1}{c|}{\textbackslash{}}                                     & \multicolumn{1}{c|}{0.010}                                                     & 0.112                                                           & \multicolumn{1}{c|}{\textbackslash{}}                                     & \multicolumn{1}{c|}{0.023}                                                     & 0.050                                                           \\ \hhline{|=|=|=|=|=|=|=|=|}
\multirow{2}{*}{\begin{tabular}[c]{@{}l@{}}Entropy\\ Reduction\end{tabular}} & \multicolumn{1}{c|}{SLMs(Ngram)} & \multicolumn{1}{c|}{\textbackslash{}}                                     & \multicolumn{1}{c|}{0.005}                                                     & 0.092                                                           & \multicolumn{1}{c|}{\textbackslash{}}                                     & \multicolumn{1}{c|}{0.010}                                                     & 0.019                                                           \\ \cline{2-8} 
                                                                             & LLMs(GPT2)                       & \multicolumn{1}{c|}{\textbackslash{}}                                     & \multicolumn{1}{c|}{0.005}                                                     & 0.042                                                           & \multicolumn{1}{c|}{\textbackslash{}}                                     & \multicolumn{1}{c|}{0.006}                                                     & 0.017                                                           \\ \hhline{|=|=|=|=|=|=|=|=|}
\multirow{4}{*}{Embedding}                                                   & SEMs(GloVe)                            & \multicolumn{1}{c|}{0.453}                                                & \multicolumn{1}{c|}{0.015}                                                     & 0.456                                                           & \multicolumn{1}{c|}{0.171}                                                & \multicolumn{1}{c|}{0.030}                                                     & 0.422                                                           \\ \cline{2-8} 
                                                                             & SEMs(W2V)                         & \multicolumn{1}{c|}{0.491}                                                & \multicolumn{1}{c|}{0.012}                                                     & 0.456                                                           & \multicolumn{1}{c|}{0.181}                                                & \multicolumn{1}{c|}{0.037}                                                     & 0.424                                                           \\ \cline{2-8} 
                                                                             & LLMs(BERT)                       & \multicolumn{1}{c|}{0.643}                                                & \multicolumn{1}{c|}{0.015}                                                     & 0.546                                                           & \multicolumn{1}{c|}{0.142}                                                & \multicolumn{1}{c|}{0.025}                                                     & 0.470                                                           \\ \cline{2-8} 
                                                                             & LLMs(GPT2)                       & \multicolumn{1}{c|}{0.639}                                                & \multicolumn{1}{c|}{0.016}                                                     & 0.561                                                           & \multicolumn{1}{c|}{0.154}                                                & \multicolumn{1}{c|}{0.028}                                                     & 0.464                                                           \\ \hline
\end{tabular}
\caption{Encoding results for three computational models on a multi-modal cognitive dataset. '/' indicates calculations are not possible due to context-based metrics, '-' signifies non-significant values, while all other numbers in the table demonstrate significant correlations with brain data.}
\end{table}

In Table 1, it is evident that, overall, large language models consistently outperform shallow embedding models in cognitive measures of embeddings, particularly in EyeTracking and fMRI-word(English). In the case of fMRI-discourse, only minor differences are observed between the performance of large and shallow language models. Concerning surprisal and entropy reduction, all models demonstrate similar levels of effectiveness. Consequently, no single model excels uniformly across all datasets; instead, each model exhibits superiority in certain aspects.

It is worth noting that existing methods often employ fMRI-discourse datasets but utilize models trained on diverse training datasets. When utilizing the same training datasets, conclusions may vary, implying that correlation results are more contingent on the training process than on the model's inherent structure. Looking ahead, future endeavors should prioritize standardizing the process of employing computational models for studying the brain. This involves releasing larger benchmark testing datasets, providing models trained on the same training dataset, employing consistent metrics for model evaluation, and sharing the code for brain encoding.

\section{Conclusion} 
Statistical language models are both simple and interpretable, explicitly encoding word co-occurrence and structure, achieving comparable performance on fMRI-discourse datasets. Shallow embedding models, which excel in learning static semantic representations, efficiently quantify meaning, presenting a novel opportunity to investigate meaning in the brain. They outperform statistical models significantly on eye-tracking datasets. Large language models, exhibiting human-like behavior, open up new possibilities for exploring hypotheses about how the brain processes language. Despite their superior performance on downstream tasks, they only outperform the other two models in the embeddings metric for fMRI-word(English) and across all cognitive metrics on the EyeTracking dataset.

This review suggests that computational models can significantly contribute to understanding the brain when used appropriately. They are particularly useful for studying cognitive load, semantic-syntactic representation, parsing strategy, and the syntactic structure of language processing in the brain. Furthermore, these models have the potential to verify existing hypotheses and generate new ones, guiding future brain studies. Despite the recent surge in attention and strong performance of large language models on downstream tasks, understanding why they correlate with the brain requires meticulous examination. It involves disentangling various factors, including training datasets, model structures, and training objectives. Conclusions drawn from such correlations are sensitive to the models used, underscoring the importance of careful consideration when extracting insights from different models. Future research should incorporate diverse datasets and various models to rigorously test the same question.

\bibliographystyle{ieeetr}
\bibliography{sample}

\end{document}